\pdfoutput=1
\documentclass[11pt]{article}

\usepackage[preprint]{acl}

\usepackage{times}
\usepackage{latexsym}
\usepackage{enumitem}

\usepackage[T1]{fontenc}
\usepackage[utf8]{inputenc}

\usepackage{microtype}

\usepackage{inconsolata}

\usepackage{graphicx}

\usepackage{amsmath}
\usepackage{amssymb}
\usepackage{amsfonts}
\usepackage{bm}
\usepackage{adjustbox}
\usepackage{multirow}
\usepackage{booktabs}
\usepackage{pifont}
\usepackage{graphicx} % For including images
\usepackage{subcaption} % For subfigures
\usepackage{algpseudocode}
\usepackage{algorithm}
\usepackage{amsmath}
\usepackage{authblk}

\newcommand{\mat}[1]{\mathbf{#1}}

\title{MaZO: Masked Zeroth-Order Optimization for Multi-Task Fine-Tuning of Large Language Models}

\author{
\vspace{-4mm}
    \textbf{Zhen Zhang\textsuperscript{1}}, 
    \textbf{Yifan Yang\textsuperscript{1}}, 
    \textbf{Kai Zhen\textsuperscript{2}}, 
    \textbf{Nathan Susanj\textsuperscript{2}}, 
    \textbf{Athanasios Mouchtaris\textsuperscript{2}},
    
    \\

    \textbf{Siegfried Kunzmann\textsuperscript{2}},
    \textbf{Zheng Zhang\textsuperscript{1}} \\
    \textsuperscript{1}University of California, Santa Barbara \\
    \textsuperscript{2}Amazon \\
    \texttt{zhen\_zhang@ucsb.edu},  \texttt{zhengzhang@ece.ucsb.edu}
}

\begin{document}
\maketitle

\begin{abstract}
Large language models have demonstrated exceptional capabilities across diverse tasks, but their fine-tuning demands significant memory, posing challenges for resource-constrained environments. Zeroth-order (ZO) optimization provides a memory-efficient alternative by eliminating the need for backpropagation. However, ZO optimization suffers from high gradient variance, and prior research has largely focused on single-task learning, leaving its application to multi-task learning unexplored. Multi-task learning is crucial for leveraging shared knowledge across tasks to improve generalization, yet it introduces unique challenges under ZO settings, such as amplified gradient variance and collinearity.
In this paper, we present MaZO, the first framework specifically designed for multi-task LLM fine-tuning under ZO optimization. MaZO tackles these challenges at the parameter level through two key innovations: a weight importance metric to identify critical parameters and a multi-task weight update mask to selectively update these parameters, reducing the dimensionality of the parameter space and mitigating task conflicts. Experiments demonstrate that MaZO achieves state-of-the-art performance, surpassing even multi-task learning methods designed for first-order optimization.

\end{abstract}

\section{Introduction}
\begin{figure*}[t]
    \centering
    \vspace{-8pt} 
    \includegraphics[width=0.8\textwidth]{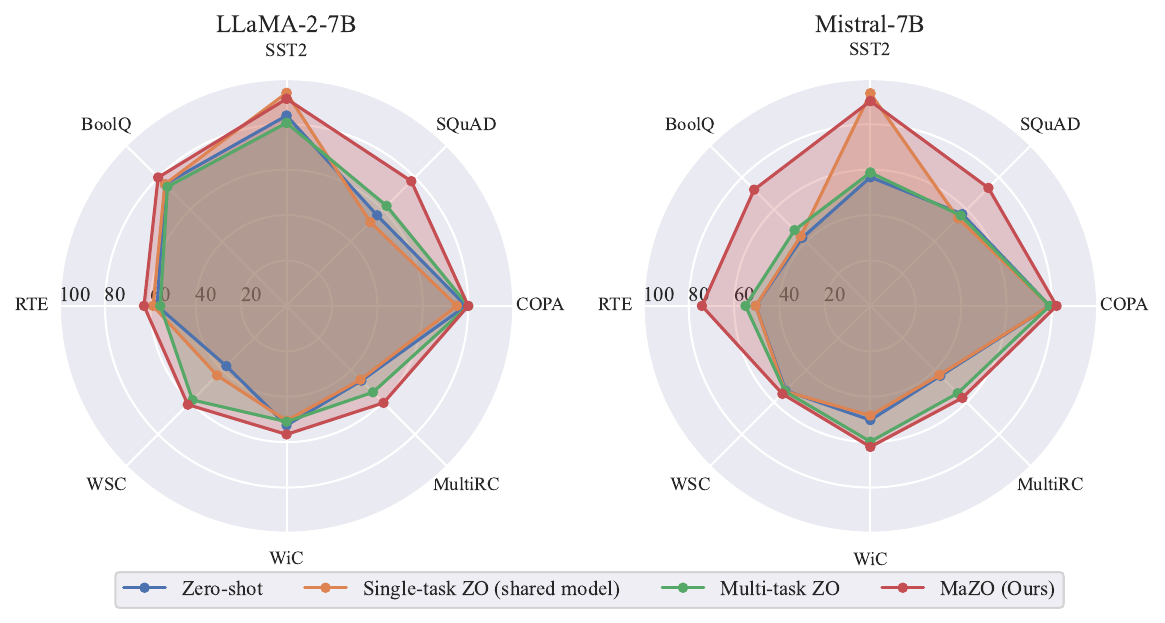}
    \vspace{-8pt} 
    \caption{Radar chart comparing the performance of our MaZO method with other methods on LLaMA-2-7B and Mistral-7B. Larger is better. Shared model means we train the model on one task and test it on all tasks.}
    \vspace{-10pt} 
    \label{fig:radar}
\end{figure*}

Large language models (LLMs) have revolutionized natural language processing, enabling breakthroughs in various applications \citep{claude35sonnet,gemini20,gpt4o,bai2023qwen}. However, the large sizes of LLMs pose significant memory challenges during training. Traditional first-order (FO) optimization uses backpropagation, which requires substantial memory to store intermediate activations and gradients \citep{rostam2024achieving, kundu2024performance}. This issue is especially pronounced in fine-tuning tasks on resource-constrained platforms (e.g. low-end GPUs or edge devices) \citep{10.1145/3666025.3699355}. Moreover, certain hardware platforms lack software support (e.g. automatic differentiation) for backpropagation \citep{bergholm2018pennylane}, further restricting FO methods. Although parameter-efficient fine-tuning methods have alleviated some of these challenges, they still require multiple times the memory of inference \citep{bai2024beyond,zhang2024revisiting}.

Zeroth-order (ZO) optimization provides a memory-efficient alternative by estimating gradients via forward passes only. Recent advances, such as MeZO \citep{malladi2023mezo}, have reduced memory usage to inference levels while achieving strong performance in LLM fine-tuning. However, the gradient variance in ZO methods is proportional to the number of perturbed parameters, which makes ZO methods struggle with high-dimensional parameter spaces, leading to slower convergence, increased gradient estimation variance, and hard to scale up \citep{chen2024enhancing}. Although recent work~\citep{liu2024sparse, yang2024adazeta,chen2023deepzero, liu2024sparse, yu2024subzero} has addressed some of these issues, most ZO methods focus on single-task learning, leaving their application to multi-task learning largely unexplored.

Multi-task learning is a key paradigm in LLMs to enable shared representations across diverse downstream tasks. This approach improves generalization, reduces the need for task-specific models, and improves performance in a wide range of applications \citep{zhang-etal-2023-survey, radford2019language}. Despite its advantages, multi-task learning also introduces inherent challenges, particularly when tasks exhibit conflicting objectives. These conflicts arise when the optimization signals from different tasks are misaligned, leading to competing gradients that prevent the model from learning effectively across all tasks~\citep{sener2018multi,Mahapatra2020MultiTaskLW,crawshaw2020multi,9382101, shi2023recon}.

The issue of conflicting gradients is further exacerbated in scenarios involving ZO optimization~\cite{liu2020primer,malladi2023mezo}. The high gradient variance in ZO methods can amplify inter-task conflicts and make it even more difficult to balance competing objectives~\citep{zhang2024convergence}. Furthermore, ZO methods suffer from collinearity in gradient estimates (see Section~\ref{sec:mtlfozo}), where aggregated gradient directions lack diversity, and higher rank in Hessian matrix (see Section~\ref{sec:challengesmtlzo}), where slower decay of eigenvalues in multi-task learning makes the convergence slow. A primary experiment demonstrated in Figure~\ref{fig:radar} shows that vanilla multi-task ZO optimization is only slightly better than zero-shot on average and is even worse on many tasks.

 % \zz{We should emphasize on the conflicts among different tasks rather than non-convexity. Single-task LLM fine-tuning is also a highly non-convex optimization problem. There is no evidence or theory to show that the proposed masking method can handle non-convexity better. }

To address these challenges, we propose Masked Zeroth-Order Optimization (MaZO), a novel framework designed for multi-task fine-tuning under ZO settings. MaZO tackles the problem at parameter level, which introduces two key innovations: (1) a weight importance metric that identifies critical parameters for each task, and (2) a multi-task weight update mask that selectively updates these parameters while freezing others. By focusing on the most important parameters, MaZO reduces the dimension of parameter space, mitigating the high variance of ZO fine-tuning while preserving the model capacity. Moreover, unlike traditional approaches dynamic weighting \citep{chen2018gradnorm,liu2024famo,AGHAJANZADEH2023109587}, which are trivial in ZO settings because of collinearity, MaZO balances multi-task learning conflicts from the perspective of weight. It activates distinct parameter subsets for different tasks based on their importance scores, allowing MaZO to allocate more capacity to tasks that require more updates.

\paragraph{Paper Contributions.} This paper makes the following novel contributions:
\begin{itemize}[leftmargin=*]
\vspace{-5pt}
    \item \textbf{First ZO-based multi-task fine-tuning framework}: We propose Masked Zeroth-Order Optimization (MaZO), the first framework specifically designed for multi-task LLM fine-tuning under ZO optimization. 
    % MaZO introduces a novel combination of weight importance metrics and a multi-task weight update mask, reducing the optimization dimensionality to lower variance while maintaining model capacity.
    \vspace{-5pt}
    \item \textbf{Task conflict resolution at the parameter level}: MaZO addresses inter-task conflicts by selectively activating critical parameters for each task. This parameter-level approach ensures balanced optimization across tasks under ZO settings.
    \vspace{-5pt}
    \item \textbf{State-of-the-art performance}: Comprehensive experiments on LLaMA-2-7B and Mistral-7B demonstrate that MaZO achieves state-of-the-art results in multi-task fine-tuning under ZO settings, outperforming multi-task learning methods designed for first-order (FO) optimization.
\end{itemize}

\section{Preliminaries and Related Work}
\label{sec:prelim_related}

\subsection{Zeroth-Order Optimization}

Zeroth-order (ZO) optimization estimates gradients using forward passes only. A common approach for ZO gradient estimation is the simultaneous perturbation stochastic approximation \citep{spall1992multivariate}, which serves as a randomized gradient estimator. Consider a model with parameters $\theta \in \mathbb{R}^d$ and a loss function $\mathcal{L}(\theta)$. Using Taylor expansion, the randomized gradient can be estimated by perturbing $\theta$ with random noise $\mathbf{z} \sim \mathcal{N}(0, \bm{I}_d)$ and computing forward and reverse losses:
\vspace{-5pt} 
\begin{align}
\widehat{\nabla} \mathcal{L}(\theta)
&= \frac{\mathcal{L}(\theta + \epsilon \mathbf{z}) - \mathcal{L}(\theta - \epsilon \mathbf{z})}{2 \epsilon} \mathbf{z},
\end{align}
\vspace{-2pt} 
\noindent where $\epsilon$ is a small scalar. The expectation of $\widehat{\nabla} \mathcal{L}(\theta)$ matches the smoothed version of the true gradient. During training, zeroth-order stochastic gradient descent (ZO-SGD) updates parameters as:
\vspace{-9pt} 
\begin{equation}
\theta = \theta - \eta \widehat{\nabla} \mathcal{L}(\theta),
\end{equation}
\vskip -9pt 
\noindent where $\eta$ is the learning rate. 

Recent advances have improved ZO optimization for large-scale applications. For example, MeZO \citep{malladi2023mezo} reduces memory usage by regenerating random perturbations $\mathbf{z}$ using random seeds instead of storing them. ZO optimization offers significant advantages for fine-tuning LLMs, as it avoids memory-intensive backpropagation \citep{liu2020primer,zhang2024revisiting}.
Despite these advantages, the gradient variance of ZO optimization increases linearly with the dimensionality of the parameter space. This leads to slower convergence and difficulties in large-scale training \citep{chen2024enhancing}. To address these challenges, various methods have been proposed. These include the design of advanced ZO optimizers \citep{zhao2024second,jiang2024zo,chen2019zo}; dimensionality reduction techniques \citep{liu2024sparse,wang2024simultaneous,yang2024adazeta,guo2024zeroth}; hybrid approaches like Addax \citep{li2024addax}; full-batch gradient estimation \citep{gautam2024variance}; exploiting low-rank structures \citep{zhao2023tensor,yu2024subzero}, and using orthogonal random directions \citep{kozak2023zeroth}.

While these methods have advanced ZO in various ways, they do not specifically address the unique challenges of multi-task learning.

\subsection{Multi-task Learning}
\label{sec:mtlfozo}
Multi-task learning aims to improve generalization performance by jointly learning $T$ related tasks through shared parameters~\citep{10.1145/3663363}. Classical multi-task learning minimizes a weighted combination of task-specific losses:
\vspace{-5pt} 
\begin{align}
\label{eq:mtl_objective}
% \small
    & \mathcal{L}(\theta) = \sum_{t=1}^T w_t \mathcal{L}^t(\theta), \text{s.t.}\ \sum_{t=1}^T w_t = 1,\ w_t \geq 0,
    % \normalsize
\end{align}
\vskip -11pt 
\noindent where $\mathcal{L}^t(\theta)$ represents the learning loss for a single task $t$. Parameter updates are performed using gradient descent. 

Multi-task learning under FO optimization has been widely studied, with different technical routes: (1) dynamic weight, which adjusts the weight of different tasks by gradients \citep{chen2018gradnorm,sener2018multi,mao-etal-2022-metaweighting}, loss \citep{liu2019loss,liu2024famo,kongyoung-etal-2020-multi,gong2024coba} or uncertainty \citep{AGHAJANZADEH2023109587}; (2) gradient manipulation \citep{desideri2012multiple,liu2021conflict,yu2020gradient}; (3) data mixing and scheduling \citep{bai2024survey,ahmadian2024mix}; (4) learning shared and specific knowledge with model architecture based on LoRA \citep{feng2024mixture,yang2024mtl,wang2023multilora} or MoE \citep{liu2023moelora,gupta2022sparsely}; (5) model merging \citep{yang2023adamerging}.

% \subsection{Challenges and Opportunities in Multi-Task Learning ZO Optimization}

% However, some of these FO methods are not directly applicable to multi-task learning ZO because the dynamic weight methods only adjust the magnitude of the gradient and pretrained model architecture is fixed.
% The collinearity issue in multi-task learning ZO highlights the need for novel approaches that operate beyond traditional strategies. While first-order multi-task learning methods provide valuable insights, multi-task learning ZO requires innovative solutions tailored to its gradient estimation framework. 
% To solve this problem, we proposed a method from the perspective of parameters instead of gradient or weights of loss from different tasks.

\section{The MaZO Framework}

\begin{figure*}[t]
    \centering
    \includegraphics[width=0.95\textwidth]{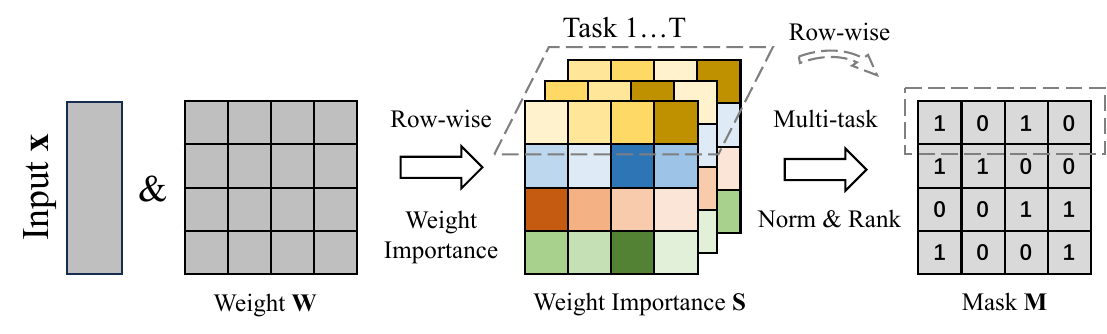} 
    \caption{Diagram of our MaZO method. The weight importance scoring and weight update mask is calculated row-wise. The weight importance for each task is calculated independently, and only from the input and weight. }
    \label{fig:mask}
    \vspace{-10pt}
\end{figure*}

\begin{figure}[t]
    \centering
    \includegraphics[width=0.48\textwidth]{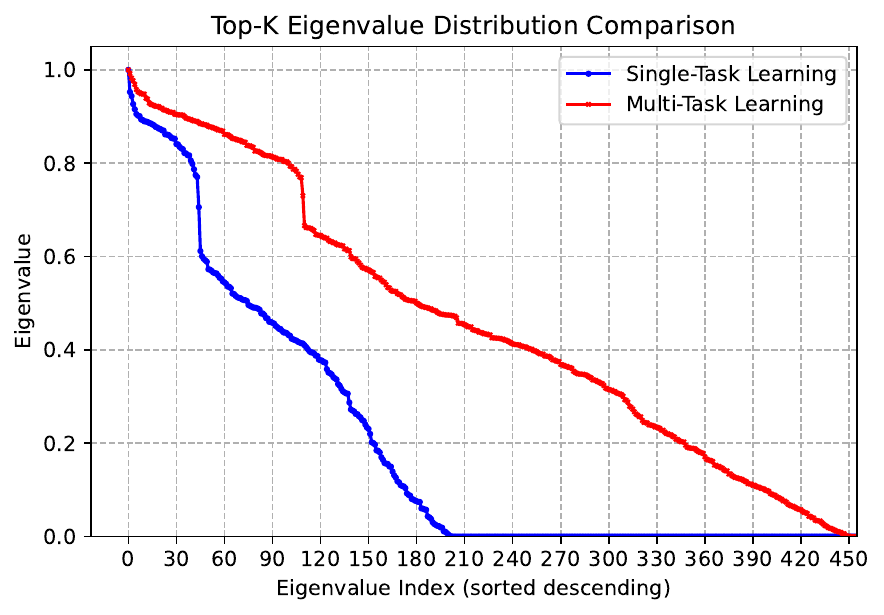} 
    \vspace{-20pt} 
    \caption{Top-K eigenvalue distribution of the Hessian matrices in multi-task learning and single-task learning. These eigenvalue are normalized by dividing by the maximum value. The slower decay of eigenvalues in multi-task learning suggests a higher effective rank, which contributes to the slower convergence of ZO fine-tuning in multi-task scenarios.}
    \vspace{-20pt} 
    \label{fig:eigenvalue_distribution}
\end{figure}

\subsection{Challenges in ZO Multi-Task Fine Tuning}
% \zz{Add a subsection to show the higher rank of the Hessian matrix in MTL. }
\label{sec:challengesmtlzo}

Under ZO optimization, multi-task learning faces unique challenges. Specifically, task-specific ZO gradient estimates exhibit fundamental collinearity, as the aggregated multi-task learning gradient aligns with the shared random perturbation $\mathbf{z}$:
\vspace{-6pt} 
\begin{align}
    \mat{g} &= \sum_{t=1}^T w_t \mat{g}^t \notag \\
    &= \left(\sum_{t=1}^T w_t \frac{\mathcal{L}^t (\theta+\epsilon \mat{z}) - \mathcal{L}^t(\theta-\epsilon \mat{z})}{2\epsilon}\right)  \mat{z}.
\end{align}
\vskip -8pt
\noindent Here $\mat{g}$ and $\mat{g}^t$ are gradients of multi-task learning and of task $t$, respectively.
This collinearity results in a lack of directional diversity, limiting optimization efficacy. Further discussion can be found in Appendix~\ref{app:collinear}.
% \zz{I'm not sure what "directional diversity" means in the mathematical sense, and why it can limit optimization efficiency.}

As explained in \citep{malladi2023mezo}, the surprising success of ZO optimization in LLM fine-tuning is due to the low-rank property of the Hessian matrix. Based on \eqref{eq:mtl_objective}, the Hessian matrix in multi-task fine-tuning can be written as
\begin{equation}
\mat{H}=\sum\limits_{t=1}^T w_t \mat{H}^t,
\end{equation}
where $\mat{H}^t$ is the Hessian associated with single-task learning loss ${\cal L}^t$. Although $\mat{H}^t$ has a low rank in the fine-tuning process, $\mat{H}$ can have a much higher rank due to the weighted sum of $T$ task-specific Hessian matrices. Figure~\ref{fig:eigenvalue_distribution} empirically verifies our theoretical claim: the Hessian in multi-task learning exhibits a broader eigenvalue spectrum than single-task learning, leading to a higher effective rank. This further slows down the convergence of ZO in multi-task LLM fine-tuning.

To address the above challenges, we propose \textbf{Masked Zeroth-Order Optimization (MaZO)}. Our approach introduces a novel framework that solves multi-task learning at parameter level. MaZO combines weight importance metrics and a multi-task weight update mask. The weight importance is derived using two complementary metrics: (1) a global score which evaluates the theoretical minimum loss when freezing a parameter, (2) a greedy score which quantifies the immediate loss change during a single optimization step. Using these scores, we construct a weight update mask that identifies a subset of critical parameters, enabling effective optimization by reducing dimensionality and variance while balancing the performance among potentially conflicting tasks. 

\subsection{Multi-Task Weight Update Mask}

We first introduce the multi-task weight update mask, assuming the weight importance scores are precomputed. We defer the computation of weight importance scores to the next subsection. In ZO optimization, the variance of an estimated gradient increases with the number of training parameters. Therefore, it is crucial to identify and focus on critical parameters for effective optimization while freezing others \citep{liu2024sparse,guo2024zeroth}.

Suppose that we have a weight importance score matrix $\mathbf{S}^t$ for each task $t$ and a sparsity level $\rho$. We unfreeze the top $k = \lceil (1 - \rho) \cdot N \rceil$ parameters in each row, where $N$ is the total number of parameters in that row. The importance scores are compared row-wise due to the approximations involved in gradient and Hessian estimation following \citet{sun2023simple}, which will be detailed in Section~\ref{sec:gradapproximation}.

Since importance scores across tasks are not directly comparable due to differing scales, we normalize the scores row-wise for each task:
\vspace{-5pt} 
\begin{equation}
\label{eq:rownorm}
\hat{\mathbf{S}}^t_{ij} = \frac{\mathbf{S}^t_{ij} - \min(\mathbf{S}^t_{i})}{\max(\mathbf{S}^t_{i}) - \min(\mathbf{S}^t_{i}) },
\end{equation}
where $\mat{S}_i^t$ denotes the $i$-th row of $\mat{S}^t$; $\hat{\mathbf{S}}^t_{ij}$ is the normalized score for parameter $j$ in row $i$ for task $t$. The overall score across tasks is computed as:
\vspace{-5pt} 
\begin{equation}
\mathbf{S} = \sum_{t=1}^T{\hat{\mathbf{S}}^t}.
\end{equation}

We select the top $k$ parameters based on $\mathbf{S}$ in {\it each row} to fine-tune, while freezing the others. This selection is represented by a binary mask matrix $\mathbf{M}$, where $\mathbf{M}_{ij} = 1$ indicates that parameter $j$ in row $i$ is unfrozen. The final parameter update is computed as:
\vspace{-8pt} 
\begin{equation}
\mathbf{\Delta W}_{\text{masked}} = \mathbf{\Delta W} \odot \mathbf{M},
\end{equation}
where $\odot$ denotes element-wise multiplication. When applied to LoRA \citep{hu2021lora}, this becomes:
\vspace{-8pt} 
\begin{equation}
\mathbf{\Delta W}_{\text{masked}} =  (\mat{A} \cdot \mat{B}) \odot \mathbf{M},
\end{equation}
where $\mat{A}$ and $\mat{B}$ are the decomposed matrices of LoRA. 

% \zz{Better to derive and use the masks for $\mat{A}$ and $\mat{B}$ respectively. It will be memory consuming if we have to reconstruct $\mathbf{\Delta W}_{\text{masked}}$ or $\mathbf{\Delta W}$ in LoRA.}

\subsection{Weight Importance}

The overall importance score for task $t$ combines the normalized global and greedy scores with a weight regularization term:
\vspace{-4pt} 
\begin{equation}
\mathbf{S}^t = \mathbf{S}^t_\text{global} +  \alpha \mathbf{S}^t_\text{greedy} + \beta |\mathbf{W}|,
\end{equation}
where $\alpha$ and $\beta$ are hyperparameters controlling the contributions of each component and $|\mathbf{W}|$ is the absolute value of weight. We now describe the computation of two complementary metrics: the \textbf{global score} and the \textbf{greedy score}.

% \zz{The weight importance scores are computed based on a scalar loss ${\cal L}$, but in multi-task learning the loss should be a vector with multiple elements. Are you talking about how to compute $\mat{S}^t$?}

\subsubsection{Global Score}

The global score is inspired by the Optimal Brain Surgeon method  \citep{frantar2023sparsegpt,sun2023simple,das2023beyond}. Unlike pruning, which sets the parameters to zero, our approach freezes certain parameters while updating others via perturbation.
Consider the Taylor expansion of the loss function of task $t$:
\begin{equation}
    \delta \mathcal{L}^t = (\mathbf{g}^t)^\top \cdot \delta \theta + \frac{1}{2} \delta \theta^\top \cdot \mathbf{H}^t \cdot \delta \theta + \mathcal{O}(\|\delta \mathbf{\theta}\|^3), \nonumber
\end{equation}
where $\mathbf{H}^t$ is the Hessian matrix of task $t$ and $\mathbf{g}^t = \frac{\partial \mathcal{L}^t_{\text{STL}}}{\partial \theta}$ . Freezing a parameter at position $m$ imposes the constraint $\mathbf{I}_m^\top \delta \theta = 0$, where $\mathbf{I}_m$ is an indicator function. The optimization problem becomes:
\begin{equation}
\begin{adjustbox}{max width=\columnwidth}
$\begin{aligned}
\min_{m} \bigg\{ 
    &\min_{\delta \theta} \bigg( 
        \left( \mathbf{g}^t \right)^\top \cdot \delta \theta 
        + \frac{1}{2} \delta \theta^\top \cdot \mathbf{H}^t \cdot \delta \theta
    \bigg) \\
    &\bigg| \mathbf{I}_m^\top \cdot \delta \mathbf{w} = 0 
\bigg\}.
\end{aligned}$
\end{adjustbox}
\end{equation}
This formulation seeks to find the parameter position $m$ that, when frozen, results in the maximal decrease in the loss function while allowing other parameters to adjust optimally. The inner optimization determines the best possible parameter updates given the constraint, while the outer optimization identifies the least impactful parameter to fix.

Using Lagrange multipliers, the optimal loss change (global score) is derived as:
\begin{align}
\label{eq:globalscore}
    (\mathbf{S}^t_\text{global})_m &=
    \delta \mathcal{L}^t_m = \frac{\left( \mathbf{I}_m^\top \cdot \left(\mathbf{H}^{t}\right)^{-1} \cdot \mathbf{g}^t \right)^2}{2 \left( \left(\mathbf{H}^t \right)^{-1} \right)_{mm}},
\end{align}
% \zz{I didn't see the dependence on $t$ on the right-hand side of the above equation.}
This expression quantifies the theoretical maximum decrease in loss when parameter $m$ is fixed, providing a measure of its importance to the overall optimization process. Smaller values indicate less important parameters, which should be frozen.

\subsubsection{Greedy Score}

Although the global score provides a theoretical measure of parameter importance, it may not suffice because the model may not converge to the optimal situation due to the large variance in the ZO gradient. Therefore, we also introduce a greedy score as a practical complement, which considers the immediate impact of freezing a parameter in a {\it single optimization step}. %This approach provides a practical complement to the global score.

For a gradient descent update with learning rate $\eta$ and random direction $\mathbf{z}$, the parameter update of task $t$ is approximated as:

\begin{equation}
\mathbf{\delta \theta} \approx - \eta \mathbf{z} \mathbf{z}^T \mathbf{g}^t.
\end{equation}
% The change in loss can be approximated using a Taylor expansion:
% \begin{align}
% \delta \mathcal{L}^t_{STL}
% &= (\mathbf{g}^t)^T \mathbf{u} + \frac{1}{2} \mathbf{u}^T \mathbf{H} \mathbf{u} + o(|\mathbf{u}|^3) \notag \\
% &= - (\mathbf{g}^t)^T \mathbf{z} \mathbf{z}^T (\mathbf{g}^t) \eta \notag \\
% &\quad + \frac{1}{2} (\mathbf{g}^t)^T \mathbf{z} \mathbf{z}^T (\mathbf{H}^t) \mathbf{z} \mathbf{z}^T (\mathbf{g}^t) \eta^2 + o(\eta^3).
% \end{align}
Substituting $\delta \theta$ and taking the expectation over random directions $z$, we obtain the expected change in loss:
\begin{align}
\mathbb{E}(\delta \mathcal{L}^t )
&= - (\mathbf{g}^t)^T \mathbf{g}^t \cdot \eta \notag \\
&+ \left( \sum_{i=0}^M (\mathbf{g}^t_i)^2 \mathbf{H}^t_{ii} + 2 (\mathbf{g}_i^t)^T \mathbf{H}^t \mathbf{g}^t \right) \eta^2 \nonumber
\end{align}
where $M$ is the number of parameters in a LLM.

When we freeze a parameter at position $m$, the change of loss (greedy score) will increase by:
\begin{align}
\label{eq:greedyscore}
(\mathbf{S}^t_\text{greedy})_m &= \delta \mathcal{L}^t_m \notag \\
&= (\mathbf{g}^t_m)^2 \eta + \mathbf{H}^t_{mm} (\mathbf{g}^t_m)^2 \eta^2 \notag \\
&\quad - 4 \sum_{j=0}^M \mathbf{H}^t_{mj} (\mathbf{g}^t_m) (\mathbf{g}^t_j) \eta^2
\end{align}
% \zz{I didn't see the dependence on $t$ on the right-hand side of the above equation.} 
Parameters with lower $\mathbf{S}^t_\text{greedy}$ values are considered less important for the current optimization step and are better candidates for freezing during multi-task learning.

% \subsection{Implementation.} 
% \label{sec:gradapproximation}

% \paragraph{Approximation of Gradient and Hessian.} In the ZO setting, directly computing the full gradient and Hessian is computationally expensive. To alleviate this, we adopt a row-wise approximation strategy following \citet{frantar2023sparsegpt}. Specifically, when considering a single row of the weight matrix, the corresponding output is a scalar, and thus the first- and second-order derivatives of the loss with respect to that output are also scalars. This observation implies that the Taylor expansion of the loss for that row leads to a gradient proportional to the input, and a Hessian proportional to the outer product of the input with itself. By dropping constant scalar factors, we approximate the gradient as $\mathbf{g}\propto \mathbf{x}$ and the Hessian as $\mathbf{H}\propto \mathbf{x}^\top\mathbf{x}$. Such an approximation not only simplifies the computation but also restricts the weight importance comparison to the row direction. Detailed derivations are provided in Appendix~\ref{app:approxiation}. \zz{Need to write this paragraph more rigorously with a few key equations in the text. }

\subsection{Implementation}  
\label{sec:gradapproximation}

To avoid the huge cost of computing the full gradient and Hessian, we adopt a row-wise approximation strategy. For a linear layer $\mathbf{y} = \mathbf{W} \mathbf{x}$, focusing on a single row \(\mathbf{w}_i\), the output is $\mathbf{y}_i = \mathbf{w}_i \mathbf{x}\). Performing a Taylor expansion of the loss \(\mathcal{L}\) with respect to \(y_i\), we find that both the first-order gradient \(\nabla \mathcal{L}(\mathbf{y}_i)\) and second-order derivative \(\nabla^2 \mathcal{L}(\mathbf{y}_i)\) are scalars. Substituting \(\Delta \mathbf{y}_i = \Delta \mathbf{w}_i \mathbf{x}\), the gradient and Hessian with respect to \(\mathbf{w}_i\) are:
\begin{align}
\mathbf{g}^t &= \frac{\partial \mathcal{L}}{\partial \mathbf{w}_i} = \nabla \mathcal{L}(\mathbf{y}_i) \mathbf{x}, \label{eq:gradient_approx} \\
\mathbf{H}^t &= \frac{\partial^2 \mathcal{L}}{\partial \mathbf{w}_i^2} = \nabla^2 \mathcal{L}(\mathbf{y}_i) (\mathbf{x} \mathbf{x}^\top). \label{eq:hessian_approx}
\end{align}
\vskip -5pt
\noindent Here, we replace the gradient with \(\mathbf{x}\), and the Hessian with \(\mathbf{x} \mathbf{x}^\top\) since we only care about the relative value {\it in a row}. This row-wise approximation significantly reduces computational cost, while still capturing the relative importance of parameters within each row. However, it also restricts the weight-importance comparison to the row direction.

\paragraph{Overall Algorithm Flow.} The pseudo-code of the whole MaZO fine-tuning framework is shown as Algorithm~\ref{alg:mazopseudocode} in Appendix~\ref{app:pseudocode}.

\section{Experiments}

\subsection{Experimental Setup}

%\textbf{Models.} 
We perform multi-task fine-tuning on two widely used decoder-only pretrained language models: LLaMA-2-7B \citep{touvron2023llama} and Mistral-7B \citep{jiang2023mistral}.% both of which are decoder-only architectures. %We use their base versions for all experiments.

\textbf{Tasks.} We evaluate our approach on a diverse set of natural language understanding (NLU) and natural language generation (NLG) tasks from the GLUE \citep{wang-etal-2018-glue} and SuperGLUE \citep{wang2019superglue} benchmarks. Specifically, for NLU, we include SST-2, BoolQ, RTE, WSC, WiC, MultiRC, and COPA, covering various classification and reasoning tasks. For NLG, we use SQuAD for question answering. To ensure computational feasibility, we sample a subset of each dataset with fixed training, validation, and test splits. Details on datasets and evaluation metrics are in Appendix~\ref{appendix:task_details}.

\textbf{Baselines.} We compare MaZO with several baselines. First, we include vanilla ZO optimization combined with traditional multi-task learning (MTL-ZO) techniques as a direct comparison to MaZO in the ZO setting. Second, we evaluate single-task learning (STL-ZO), where models are trained individually on each task to provide an {\bf upper bound} for task-specific performance without multi-task conflicts, as well as a single-task transfer baseline, where the model is trained on a single task (SST-2) using vanilla ZO optimization and evaluated across all tasks to highlight the limitations of single-task training in multi-task scenarios. Third, we include LoRA fine-tuning \citep{hu2021lora}, a parameter-efficient approach, and extend MaZO to update LoRA matrices under ZO settings, demonstrating its flexibility. Finally, we compare MaZO against state-of-the-art first-order (FO) multi-task learning methods, including CoBa \citep{gong2024coba}, FAMO \citep{liu2024famo}, and MTL-LoRA \citep{yang2024mtl}, to its compatibility with ZO optimization. These baselines provide a comprehensive comparison for assessing MaZO's effectiveness and robustness in addressing the challenges of ZO-based multi-task learning.

% \zz{Better to itemize these baselines and provide very short descriptions for each baseline method.}

\begin{table*}[t]
\centering
\resizebox{\textwidth}{!}{
\begin{tabular}{lcccccccccc}
\toprule
\textbf{Task} & \textbf{SST-2} & \textbf{BoolQ} & \textbf{RTE} & \textbf{WSC} & \textbf{WiC} & \textbf{MultiRC} & \textbf{COPA} & \textbf{SQuAD} & \multirow{2}{*}{\textbf{Avg}} \\ 
\textbf{Task Type} & \multicolumn{6}{c}{---------------------- \textit{Classification} ----------------------} & \multicolumn{1}{c}{-- \textit{Multiple Choice} --} & \multicolumn{1}{c}{-- \textit{Generation} --}\\
\midrule
$\text{STL-ZO}$ (one model per task)   &  93.8           & 83.0           & 73.5         & 51.3         & 62.1         & 61.0            & 86.0          & 79.6          & 73.8         \\
Zero-Shot         &  83.8           & 75.8           & 57.0         & 37.5         & 52.6         & 46.6            & 79.0          & 56.4          & 61.1         \\
ICL               &  93.7           & 78.7           & 61.2         & 47.2         & 59.9         & 54.3            & 80          & 57.7          & 66.6         \\
$\text{STL-ZO} (\text{shared model}) $    & 93.8           & 75.8           & 58.8         & 43.3         & 50.6         & 46.0            & 75.0          & 52.2          & 61.9         \\
$\text{MTL-ZO}$     & 80.6           & 74.2           & 55.6         & 58.6         & 51.0         & 53.8            & \textbf{80.0}          & 62.3          & 64.5         \\
$\text{MTL-ZO}_{\text{LoRA}}$           & 90.2           & \textbf{80.0}           & 61.0         & 54.8         & 56.6         & 58.0            & 76.0          & 74.8          & 68.9         \\
\midrule
$\text{MTL-ZO}_{\text{MTL-LoRA}}$        & 88.4           & 76.8           & 60.2         & 60.6         & \textbf{57.6}         & \textbf{61.6}            & 79.0          & 59.4          & 68.0         \\
$\text{MTL-ZO}_{\text{CoBa}}$            & 82.8           & 75.3           & 56.8         & 60.6         & 53.4         & 55.8            & 77.0          & 57.6          & 64.9         \\
$\text{MTL-ZO}_{\text{FAMO}}$            & 91.0           & 77.6           & 59.4         & 56.5         & 53.4         & 50.6            & 78.0          & 53.9          & 65.1         \\
\midrule
\textbf{MaZO}     &  90.6  & 76  & \textbf{62.8} & 56.7 & 52.6 & 58.6   & \textbf{82.0} & 55.5 & 66.9 \\ 
\textbf{$\text{MaZO}_{\text{LoRA}}$}     &  \textbf{91.2}  & \textbf{80.0}  & \textbf{62.8} & \textbf{61.5} & 56.6 & 60.4   & 80.0 & \textbf{77.7} & \textbf{71.3} \\ 
\bottomrule
\end{tabular}}
\caption{Performance comparison across tasks using different methods on LLaMA-2-7B. The average score (Avg) is computed across all tasks. Metrics for these tasks are consistent with MeZO \citep{malladi2023mezo}. \textit{Shared model} indicates training on a single task (SST-2) and testing on all tasks. \textit{ICL} refers to in-context learning. \textit{STL} represents single-task learning and \textit{MTL} represents multi-task learning.}

\label{tab:performance_comparison}
\end{table*}

\begin{figure}[t]
    \centering
    
    \includegraphics[width=0.48\textwidth]{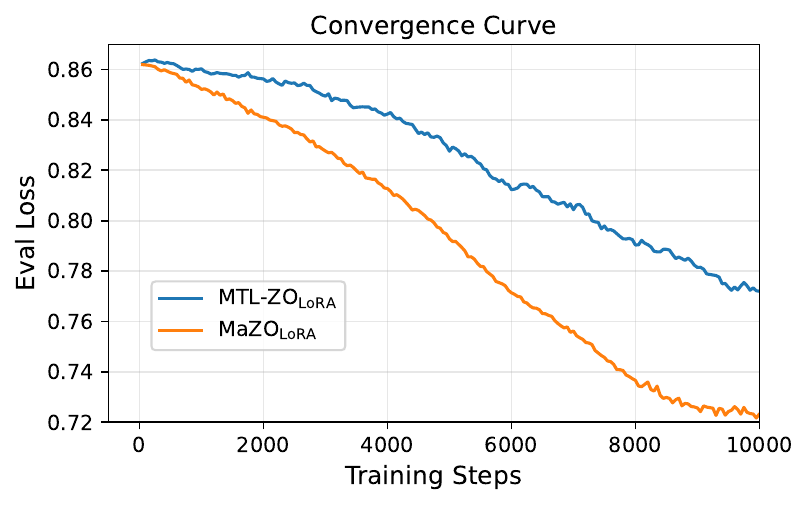} 
    % \vspace{-20pt} 
    \caption{The convergence curve of (1) vanilla multi-task ZO fine-tuning with LoRA, (2) MaZO with LoRA. }
    % \vspace{-10pt} 
    \vspace{-10pt}
    \label{fig:convergence}
\end{figure} 

\subsection{Results on LLaMA-2-7B}
\paragraph{MaZO Outperforms Competitors.} 
The results for LLaMA-2-7B are presented in Table~\ref{tab:performance_comparison}. Vanilla multi-task ZO optimization shows only slight improvements over the zero-shot baseline, highlighting its inability to effectively address multi-task conflicts under ZO settings. Similarly, vanilla single-task ZO optimization with a shared model fails to generalize effectively across multiple tasks, underscoring the inherent challenges of ZO optimization in multi-task scenarios.

In contrast, our proposed MaZO framework achieves the {\bf highest average performance} across all tasks and demonstrates a {\bf balanced} performance profile. These results validate MaZO's ability to mitigate inter-task conflicts and optimize multi-task learning by selectively focusing on critical parameters. The effectiveness of MaZO is further evident in its superior performance in both full-model ZO fine-tuning and LoRA-based fine-tuning, with particularly pronounced gains in the latter. This underscores MaZO's flexibility and its compatibility with parameter-efficient fine-tuning techniques.

\paragraph{Dimensionality Reduction Enhances Multi-Task Learning.} 
The application of LoRA to ZO fine-tuning significantly improves the performance of multi-task learning. This improvement can be attributed to LoRA's ability to reduce the dimensionality of the parameter space, thereby lowering the variance of gradient estimates. These findings reinforce the validity of MaZO's masking strategy, which optimizes multi-task learning by focusing on a reduced set of critical parameters.

\paragraph{FO multi-task learning methods do not apply to ZO.} Multi-task learning methods originally developed for first-order (FO) optimization, such as CoBa and FAMO, do not achieve effective performance in the ZO setting. This can be attributed to their inability to resolve multi-task conflicts due to the collinearity problem in ZO gradient estimates. Under the ZO framework, FO methods can only adjust the magnitude of the approximated gradient, but not its direction, resulting in  performance degradation. 
Additionally, MTL-LoRA, the multi-task version of LoRA fine-tuning, does not significantly enhance performance in the ZO setting. This may be due to the sensitivity of task-specific weights and the diagonal transformation matrix to noise. Perturbation-based optimization, as used in ZO, introduces excessive variance, which undermines the effectiveness of these FO-based methods.

\begin{table*}[t]
\centering
\resizebox{\textwidth}{!}{
\begin{tabular}{lcccccccccc}
\toprule
\textbf{Task} & \textbf{SST-2} & \textbf{BoolQ} & \textbf{RTE} & \textbf{WSC} & \textbf{WiC} & \textbf{MultiRC} & \textbf{COPA} & \textbf{SQuAD} & \multirow{2}{*}{\textbf{Avg}} \\ 
\textbf{Task Type} & \multicolumn{6}{c}{---------------------- \textit{Classification} ----------------------} & \multicolumn{1}{c}{-- \textit{Multiple Choice} --} & \multicolumn{1}{c}{-- \textit{Generation} --}\\
\midrule
$\text{STL-ZO}$ (one model per task)       & 93.6           & 77.8           & 74.2         & 55.3         & 62.1         & 62.7            & 88.0          & 76.5          & 73.8         \\
Zero-Shot         & 56.7           & 42.4           & 50.5         & 52.8         & 50.3         & 43.6            & 79.0          & 57.2          & 54.1         \\
ICL               & 62.3           & 46.1           & 56.0         & 53.2         & 61.4         & 53.4            & 79.0          & 62.3          & 59.2         \\
$\text{MTL-ZO}$      &  58.7           & 47.2           & 55.0         & 53.2         & 59.8         & 54.4            & 79.0          & 56.3          & 58.0         \\
$\text{MTL-ZO}_{\text{LoRA}}$           & 89.3           & \textbf{73.2}           & 71.5         & 51.3         & 58.1         & 53.4            & 80.0          & \textbf{73.5}          & 68.7         \\
\midrule
\textbf{MaZO}     & 83.4  & 56.3  & 60.2 & 54.8 & 58.1 & 55.8 & 79 & 59.4 & 63.4 \\ 
\textbf{$\text{MaZO}_{\text{LoRA}}$}    & \textbf{90.2}  & 72.4  & \textbf{74.2} & \textbf{54.8} & \textbf{62.1} & \textbf{57.3}   & \textbf{82.0} & \textbf{73.5} & \textbf{70.8} \\ 
\bottomrule
\end{tabular}}
\caption{Performance comparison across tasks using different methods on Mistral-7B. The setting and notation are the same as Table~\ref{tab:performance_comparison}. We exclude the FO MTL methods as they do not have significant improvement.}
\label{tab:mistral_performance_comparison}
% \vspace{-10pt}
\end{table*}
% \vskip 8em

\subsection{Results on Mistral-7B}  
The results for Mistral-7B in Table~\ref{tab:mistral_performance_comparison} reveal trends similar to those observed with LLaMA-2-7B. Despite the relatively low zero-shot performance of Mistral-7B, vanilla multi-task learning ZO fails to deliver substantial improvements. This underscores the inherent challenges of ZO-based multi-task learning.
In contrast, MaZO consistently outperforms all other methods. Its ability to mitigate ZO-specific challenges is evident in its superior performance, further validating MaZO as a state-of-the-art solution for ZO-based multi-task learning.

\subsection{Computational Performance}
  
% \zz{The training diverged in the later stage, which is very bad.. We may need some variance reduction approaches in the later stage to improve the convergence. I'm not sure if this doable before the ACL deadline. If not, we may remove the convergence curve, keep working on this and show an improved convergence curve in the rebuttal.}
Figure~\ref{fig:convergence} shows that MaZO converges faster and achieves a significantly lower loss compared to traditional multi-task ZO fine-tuning methods. This holds true both with and without LoRA. This improvement can be attributed to the mask mechanism in MaZO, which focuses on optimizing the most critical parameters, thereby reducing gradient noise, balancing the inter-task conflicts, and accelerating convergence.

The significantly better convergence and accuracy of MaZO is obtained at marginal computing and memory overhead. Specifically, the mask search time introduced by MaZO is negligible compared to the overall training time. When evaluated in the LlaMMA-7B model, MaZO incurs a slight increase in memory usage (approximately $10\%$) compared to baseline multi-task learning ZO methods. This is primarily due to the additional storage required for the weight update mask. However, this increase does not significantly impact the overall memory efficiency, especially when combined with LoRA, where the parameter space is already reduced. Details are provided in Appendix~\ref{sec:memory usage}.
%While MaZO introduces a small memory overhead, its benefits in terms of faster convergence and reduced gradient variance outweigh this cost, making it an effective and practical solution for multi-task fine-tuning under ZO optimization.

\subsection{Various Weight Importance Metrics}  

% \zz{Compare with standard sparse ZO fine-tuning.}
\begin{table}[t]
    \centering
    \resizebox{0.5\textwidth}{!}{
    \begin{tabular}{l|ccccc}
        \toprule
        \textbf{Task} & \textbf{SST-2} & \textbf{BoolQ} & \textbf{Copa} & \textbf{SQuAD} & \textbf{Avg}\\ 
        \midrule
        No Mask    &  85.4 & 72.2 & 80.0 & 66.0 & 75.9 \\
        Random     &  86.6 & 73.0 & 80.0 & 63.4 & 75.8 \\
        Magnitude  &  87.4 & 75.6 & 79.0 & 65.6 & 76.9 \\
        Wanda      &  88.4 & 77.8 & 80.0 & 62.4 & 77.2 \\
        MaZO       &  90.2 & 78.0 & 81.0 & 72.3 & 80.4 \\
        \bottomrule
    \end{tabular}}
    \caption{Comparison of different weight importance metrics. The sparsity is set to 50\% except for \textit{No Mask}. Random and Magnitude are done weight-wise while Wanda and MaZO are selected row-wise.}
    \label{tab:maskmetric}
    \vspace{-8pt} 
\end{table}

To further validate the effectiveness of MaZO, we compare its performance with three alternative weight scoring methods: random selection, magnitude-based scoring, and Wanda scoring. Detailed implementation of these methods is described in Appendix~\ref{app:metrics}. For a fair comparison, we fix the sparsity level at 50\%, consistent with the sparsity used in the Wanda score. Table~\ref{tab:maskmetric} summarizes the results of this comparison.

The findings indicate that while both the magnitude-based and Wanda-based scoring can improve average performance, their improvements are less pronounced and less balanced across tasks compared to MaZO. This is because these methods evaluate the weight importance statically, without considering training dynamics or perturbation-based insights. In contrast, MaZO dynamically identifies critical parameters during training, enabling more effective optimization and better multi-task balance under the ZO framework. These results underscore the superiority of MaZO in leveraging weight importance to achieve state-of-the-art performance in multi-task fine-tuning.

\subsection{Ablation Study}  
%\zz{Reduce this subsection by about a half.}
%\subsubsection{Hyperparameters}  

Finally, we explore the optimal hyperparameter settings for MaZO, includeing $\alpha$, $\beta$, sparsity level, and the LoRA rank. To streamline the process, we perform grid searches for each hyperparameter while keeping the others constant. For most experiments, we fine-tune the model on SST-2, BoolQ, COPA, and SQuAD, encompassing binary classification, multiple-choice, and generation tasks, providing diverse evaluation scenarios. However, for the LoRA rank, we evaluate performance across all tasks.

\textbf{$\alpha$ and $\beta$.}  
To optimize $\alpha$ and $\beta$, we fix the sparsity level at 50\% and perform full-model fine-tuning (without LoRA). The search is conducted in two stages. First, $\beta$ is set to zero, and $\alpha$ is tuned, resulting in an optimal value of $\alpha = 10$. Next, with $\alpha$ fixed, $\beta$ is tuned, yielding an optimal value of $\beta = 1$. These values strike a balance between the global and greedy weight importance metrics, ensuring effective parameter selection.

\textbf{LoRA rank.} 
We examine the impact of LoRA rank and provide detailed results in Appendix~\ref{app:lorarank}. In summary, the results reveal a U-shaped relationship between rank and performance, reflecting a trade-off between model capacity and dimensionality. The optimal rank of $16$ minimizes loss and is used as the default setting for LoRA-based baseline.

\textbf{Sparsity.}  
We perform a grid search of the sparsity level $\rho$ from $0.1$ to $0.99$.
For full-model fine-tuning, the performance first improves with increasing sparsity and then sharply declines. The peak performance is achieved at $\rho = 0.9$. For LoRA fine-tuning, we jointly optimize sparsity levels and LoRA ranks. %The LoRA rank is searched from 16 to 512. 
The optimal result is found at a LoRA rank of 64 and a sparsity level of $0.8$. Notably, the effective number of parameters is equivalent to $64 \times (1-0.8) = 12.8$, which is less than the best-performing rank of LoRA baseline. This highlights that MaZO can further reduce the dimension while maintaining the model capacity.

% \section{Discussion}
\vspace{-1pt}
\section{Conclusion}
\vspace{-1pt}
In this work, we have presented MaZO, a novel framework that harnesses masked zeroth-order optimization for the multi-task fine-tuning of LLMs. By incorporating weight importance score alongside a multi-task weight update mask, MaZO effectively reduces gradient variance and mitigates conflicts among tasks. Our experimental results demonstrate that MaZO not only surpasses current zeroth-order optimization methods but also outperforms leading multi-task learning methods designed for first-order optimization across a range of NLP tasks. Furthermore, our parameter-level approach is not limited solely to zeroth-order optimization, offering potential integrations with a variety of other optimization strategies.

\section{Limitations}
While MaZO demonstrates strong empirical performance, several limitations warrant discussion. First, the computation of weight importance introduces additional computational overhead compared to vanilla ZO methods. However, this cost remains negligible relative to the memory and computational demands of model weights and activations. Second, the effectiveness of MaZO is partially contingent on the quality of gradient and Hessian approximations. While our current approximations are effective, they could be further refined through more sophisticated estimation techniques to enhance performance. Third, our experiments are limited to medium-scale models (7B parameters) due to computational constraints. Although the method is theoretically applicable to larger models, the interplay between mask sparsity and model scale has not been systematically studied and represents an avenue for future research. Finally, we do not provide a theoretical convergence analysis for the masking approach. However, Sparse MeZO~\citep{liu2024sparse} has already conducted a comprehensive and rigorous analysis of general masking scenarios in zeroth-order optimization. We refer interested readers to their work for detailed theoretical insights, and therefore do not duplicate these efforts here.

\bibliography{acl_latex}

\newpage
\appendix

\section{Additional Explanation on Hessian and Gradient Approximation}
\label{app:approxiation}

Consider a linear layer in an LLM that computes:
\begin{equation}
\mathbf{y} = \mathbf{W} \mathbf{x},
\end{equation}
where \(\mathbf{W} \in \mathbb{R}^{m \times n}\), \(\mathbf{x} \in \mathbb{R}^{n}\), and \(\mathbf{y} \in \mathbb{R}^{m}\). Focusing on one particular linear component, let us analyze a single row \(\mathbf{w}_i \in \mathbb{R}^{n}\) of \(\mathbf{W}\). The corresponding output is given by:
\begin{equation}
\mathbf{y}_i = \mathbf{w}_i \, \mathbf{x},
\end{equation}
which is a scalar.

To analyze the sensitivity of the loss \(\mathcal{L}\) with respect to \(\mathbf{w}_i\), we perform a second-order Taylor expansion of \(\mathcal{L}\) with respect to \(y_i\):
\begin{align}
\mathcal{L}(\mathbf{y}_i + \Delta \mathbf{y}_i) &\approx \mathcal{L}(\mathbf{y}_i) + \nabla \mathcal{L}(\mathbf{y}_i)\,\Delta \mathbf{y}_i \notag \\
&\quad + \frac{1}{2}\nabla^2 \mathcal{L}(\mathbf{y}_i)\,(\Delta \mathbf{y}_i)^2.
\end{align}
Since \(\mathbf{y}_i\) is a scalar, its second derivative \(\nabla^2 \mathcal{L}(\mathbf{y}_i)\) is also a scalar.

Now, the change in \(\mathbf{y}_i\) due to a change in the weights is
\begin{equation}
\Delta \mathbf{y}_i = \Delta \mathbf{w}_i\, \mathbf{x}.
\end{equation}
Substituting this into the second-order term yields:
\begin{equation}
\frac{\partial^2 \mathcal{L}}{\partial \mathbf{w}_i^2} \approx \nabla^2 \mathcal{L}(\mathbf{y}_i)\, (\mathbf{x} \mathbf{x}^\top).
\end{equation}

Since we are primarily interested in comparing weight importance along the row direction, the absolute scale of the Hessian is not crucial. In practice, we can drop the multiplicative factor \(\nabla^2 \mathcal{L}(y_i)\) (or, equivalently, assume it to be a constant) and write:
\begin{equation}
\frac{\partial^2 \mathcal{L}}{\partial \mathbf{w}_i^2} \propto \mathbf{x} \mathbf{x}^\top.
\end{equation}

Similarly, one can derive a first-order approximation for the gradient. By retaining only the first-order term of the Taylor expansion, we have:
\begin{equation}
\mathcal{L}(\mathbf{y}_i + \Delta \mathbf{y}_i) \approx \mathcal{L}(\mathbf{y}_i) + \nabla \mathcal{L}(\mathbf{y}_i)\, \Delta \mathbf{y}_i.
\end{equation}
With \(\Delta \mathbf{y}_i = \Delta \mathbf{w}_i \, \mathbf{x}\), the gradient with respect to \(\mathbf{w}_i\) becomes:
\begin{equation}
\frac{\partial \mathcal{L}}{\partial \mathbf{w}_i} \approx \nabla \mathcal{L}(\mathbf{y}_i)\, \mathbf{x}.
\end{equation}
Similarly, since we are only interested in the relative value, the factor is dropped:
\begin{equation}
\mathbf{g} \propto \mathbf{x}.
\end{equation}

This derivation shows that, by considering each row independently (row-wise), we avoid the immense complexity involved in computing the full Hessian matrix (which is high-dimensional and difficult to characterize even under diagonalization assumptions). In other words, computing the Hessian row-wise allows us to circumvent the problem of determining the eigenvalues or even a reliable diagonal approximation of the full Hessian.

%\label{mazopseudocode}
\begin{algorithm}[t!]
\caption{MaZO LLM Fine-Tuning Framework}
\label{alg:mazopseudocode}
\begin{algorithmic}[t]
\State \textbf{Input:} 
     Pre-trained LLM parameters $\theta$, training data, tasks $t = 1, \dots, T$, 
     sparsity level $\rho$, hyperparameters $\alpha$, $\beta$, learning rate $\eta$
\State \textbf{Output:} Updated parameters $\theta^*$

\For{each task $t=1$ to $T$}
    \State Collect evaluation data
    \State \textbf{Compute} $\mathbf{S}^t_{\text{global}}$ with eq. (\ref{eq:globalscore})
    \State \textbf{Compute} $\mathbf{S}^t_{\text{greedy}}$ with eq. (\ref{eq:greedyscore})
    \State \textbf{Combine} scores:
        \[
        \mathbf{S}^t = \mathbf{S}^t_{\text{global}} + \alpha\,\mathbf{S}^t_{\text{greedy}} + \beta|\mathbf{W}|
        \]
    \State \textbf{Normalize} $\mathbf{S}^t$ and get $\hat{\mathbf{S}}^t$ with eq. (\ref{eq:rownorm})
\EndFor

\State \textbf{Aggregate:} Sum row-wise normalized scores across tasks:
    \[
    \mathbf{S} = \sum_{t=1}^{T}\hat{\mathbf{S}}^t
    \]
\For{each weight $\mathbf{W}$ in LLM}
\For{each row $i$ in $\mathbf{W}$}
    \State Select top $k$ parameters in each row according to corresponding $\mathbf{S}$ and construct weight update mask $\mathbf{M}$ 
\EndFor
\EndFor
\For{each training step}
\State \textbf{Compute} weight update $\Delta \mathbf{W}$ using ZO optimization
\State \textbf{Apply mask:} 
    \[
    \Delta \mathbf{W}_{\text{masked}} = \Delta \mathbf{W} \odot \mathbf{M}
    \]
\State \textbf{Update} parameters: 
    \[
    \theta \leftarrow \theta + \eta \Delta \mathbf{W}_{\text{masked}}
    \]
\EndFor
\end{algorithmic}
\end{algorithm}

\section{Pseudo-code of MaZO}
\label{app:pseudocode}
The pseudo-code of the whole MaZO LLM fine-tuning framework is shown as Algorithm~\ref{alg:mazopseudocode}.

\section{Baseline}
\label{app:baseline}

\subsection{CoBa: Convergence Balancer for Multitask Finetuning}
CoBa (Convergence Balancer) \citep{gong2024coba} is a novel multi-task learning (MTL) method designed for large language models (LLMs). It dynamically adjusts task weights during training to ensure balanced convergence across tasks, while maintaining computational efficiency. 

Consider an LLM parameterized by $\theta \in \mathbb{R}^m$, trained on $T \geq 2$ tasks. The loss function for task $t$ at iteration $i$ is denoted as $\mathcal{L}^t(\theta; i): \mathbb{R}^m \to \mathbb{R}_{\geq 0}$. The overall optimization objective is:
\begin{equation}
\min_{\theta \in \mathbb{R}^m} \mathcal{L}(\theta; i) = \sum_{t=1}^T \omega_t(i) \mathcal{L}^t(\theta; i),
\end{equation}
where $\omega_t(i)$ is the weight of task $t$ at iteration $i$. A uniform weight assignment $\omega_t(i) = \frac{1}{T}$ ensures equal attention to all tasks but often leads to varying convergence rates. CoBa dynamically adjusts $\omega_t(i)$ to balance these rates, prioritizing generalization by deriving weights from validation losses instead of training losses.
CoBa is built upon three main components:

\textbf{Relative Convergence Score (RCS)} dynamically allocates smaller weights to tasks that converge faster and larger weights to slower-converging tasks. It is computed as:
\begin{equation}
\label{eq:rcs}
\text{RCS}_t(i) = \text{softmax}_t \left( T \frac{\alpha_t(i)}{\sum_{t'=1}^T |\alpha_{t'}(i)|} \right),
\end{equation}
where $\alpha_t(i)$ is the convergence slope of task $t$, derived from the normalized validation loss ratio over a sliding window of $N$ iterations. The softmax operation ensures differentiation across tasks, with faster-converging tasks receiving lower weights.

\textbf{Absolute Convergence Score (ACS)} addresses task divergence by reducing weights for diverging tasks and increasing weights for converging tasks. It is computed as:
\begin{equation}
\label{eq:acs}
\text{ACS}_t(i) = \text{softmax}_t \left( -N \frac{\alpha_t(i)}{\sum_{j=i-N+1}^i |\alpha_t(j)|} \right),
\end{equation}
where normalization is performed along the historical iteration dimension, isolating a task's own trajectory. ACS ensures tasks with consistent convergence receive higher weights while diverging tasks are penalized.

\textbf{Divergence Factor (DF)} determines the relative influence of RCS and ACS on the final task weights. It is defined as:
\begin{equation}
\label{eq:df}
\text{DF}(i) = \min \left( \text{softmax}_i \left( \frac{i \cdot \alpha_{\text{max}}(i)}{\sum_{j=1}^i \alpha_{\text{max}}(j)} \right), 1 \right),
\end{equation}
where $\alpha_{\text{max}}(i)$ is the largest convergence slope across all tasks at iteration $i$. DF ensures RCS dominates when all tasks are converging, while ACS takes precedence when divergence is detected.

\textbf{The final task weights} $\omega_t(i)$ are computed as:
\begin{equation}
\label{eq:cobaw}
\omega_t(i) = \text{DF}(i) \cdot \text{RCS}_t(i) + (1 - \text{DF}(i)) \cdot \text{ACS}_t(i),
\end{equation}
allowing a seamless transition between RCS and ACS dominance based on task convergence trends.

\textbf{The convergence slope} $\alpha_t(i)$ for task $t$ is calculated based on the normalized validation loss ratio $\bar{\mathcal{L}}_t^{\text{val}}(\theta; i)$. Specifically, we fit a linear model to the validation loss ratios over a sliding window of $N$ iterations. The observations are defined as:
\begin{equation}
\mathbf{x}_t(i) = [i, 1]^\top, \quad \mathbf{X}_t(N; i) = [\mathbf{x}_t(s_0), \dots, \mathbf{x}_t(i)]^\top,
\end{equation}
\begin{equation}
\mathbf{y}_t(N; i) = [\bar{\mathcal{L}}_t^{\text{val}}(\theta; s_0), \dots, \bar{\mathcal{L}}_t^{\text{val}}(\theta; i)]^\top,
\end{equation}
where $s_0 = \max(0, i - N + 1)$ is the starting step of the sliding window. The goal is to compute the coefficient vector $c_t(N; i) = [\alpha_t(N; i), \beta_t(N; i)]^\top$ that minimizes the mean squared error (MSE) between the predicted and actual validation loss ratios:
\begin{equation}
c_t = \arg \min_{c_t} \frac{1}{2} (\mathbf{X}_t c_t - \mathbf{y}_t)^\top (\mathbf{X}_t c_t - \mathbf{y}_t).
\end{equation}

The closed-form solution for $c_t$ is given by:
\begin{equation}
c_t = (\mathbf{X}_t^\top \mathbf{X}_t)^{-1} \mathbf{X}_t^\top \mathbf{y}_t.
\end{equation}

\paragraph{Algorithm}
The CoBa algorithm is summarized in Algorithm~\ref{alg:coba}, We use $M = 4$ with $\text{batchsize}=16$

\begin{algorithm}[t]
\caption{CoBa Algorithm}
\label{alg:coba}
\begin{algorithmic}[1]
\Require Initial parameters $\theta_0$, $M$ batches of validation set, history window length $N = 5M$, warm-up steps $W = M$, number of tasks $T$, initial weights $\omega_i(0) = \frac{1}{T}$.
\Ensure Trained parameters $\theta$.
\For{$t = 0$ to $T$}
    \State Compute $\mathcal{L}(\theta; i)$ with training batch $x_i$.
    \State Compute $\bar{\mathcal{L}}_t^{\text{val}}(\theta; t)$ with validation batch $v_i$.
    \State Update validation loss history $\mathbf{y}_t(N; i)$.
    \State Compute $\alpha_t(i)$.
    \If{$i > W$}
        \State Compute $\text{RCS}(i)$, $\text{ACS}(i)$, and $\text{DF}(i)$ using Eqs. (\ref{eq:rcs}), (\ref{eq:acs}), and (\ref{eq:df}).
        \State Update task weights $\omega_t(i)$ using Eq. (\ref{eq:cobaw}).
    \Else
        \State Set $\omega_t(i) = \frac{1}{T}$.
    \EndIf
\State Update model parameters $\theta$ using weighted loss $\mathcal{L}(\theta; i)$.
\EndFor
\end{algorithmic}
\end{algorithm}

\subsection{FAMO: Fast Adaptive Multitask Optimization}

Fast Adaptive Multitask Optimization (FAMO) is a dynamic weighting method designed to address the challenges of multitask learning (MTL), where directly optimizing the average loss across tasks often leads to under-optimization of certain tasks. FAMO ensures balanced task loss reduction using only $O(1)$ space and time per iteration, making it computationally efficient and scalable.

The complete FAMO algorithm is summarized in Algorithm~\ref{alg:famo}.

\begin{algorithm}[t]
\caption{PyTorch Implementation of Wanda}
\label{alg:wanda}
\begin{algorithmic}
\State \textbf{Input:} Weight matrix $\mathbf{W} \in \mathbb{R}^{C_{\text{out}} \times C_{\text{in}}}$, input activations $\mathbf{X} \in \mathbb{R}^{(N \cdot L) \times C_{\text{in}}}$, sparsity ratio $s \in [0, 1]$
\State \textbf{Output:} Pruned weight matrix $\mathbf{W}$
\State Compute importance scores: $\text{metric} = \mathbf{W}.\text{abs()} \cdot \mathbf{X}.\text{norm}(p=2, \text{dim}=0)$
\State Sort scores \textbf{within each row}: $\_, \text{sorted\_idx} = \text{torch.sort}(\text{metric}, \text{dim}=1)$
\State Identify indices to prune: $\text{pruned\_idx} = \text{sorted\_idx}[:, :\lfloor C_{\text{in}} \cdot s \rfloor]$
\State Set pruned weights to zero: $\mathbf{W}.\text{scatter\_}(\text{dim}=1, \text{index}=\text{pruned\_idx}, \text{src}=0)$
\State \textbf{Return} $\mathbf{W}$
\end{algorithmic}
\end{algorithm}

\begin{algorithm}[t]
\caption{Fast Adaptive Multitask Optimization (FAMO)}
\label{alg:famo}
\begin{algorithmic}[1]
\Require Initial model parameters $\theta_0$, task losses $\{\mathcal{L}_{t,i}\}_{t=1}^T$, learning rates $\alpha$ and $\beta$, decay factor $\gamma$.
\State Initialize logits: $\mathbf{\xi}_1 \gets \mathbf{0}$.
\For{$i = 1$ to $T$}
    \State Compute task weights: 
    \[
    \mathbf{z}_i = \operatorname{Softmax}(\mathbf{\xi}_t),
    \]
    where for each $i$, 
    \[
    \mathbf{z}_{t,i} = \frac{\exp(\mathbf{\xi}_{t,i})}{\sum_{t'=1}^T \exp(\mathbf{\xi}_{t',i})}.
    \]
    \State Update model parameters:
    \[
    \theta_{t+1} = \theta_t - \alpha\,  \sum_{t=1}^T \left( c_t \frac{\mathbf{z}_{t,i}}{\mathcal{L}_{t,i}}\,\right) \nabla \mathcal{L}_{t,i},
    \]
    \[
    \text{with} \quad
    c_i = \left(\sum_{i=1}^k \frac{\mathbf{z}_{t,i}}{\mathcal{L}_{t,i}} \right)^{-1}.
    \]
    \State Compute the vector of log-loss differences:
    \[
    d_i = \begin{bmatrix}
      \log \mathcal{L}_{1,i} - \log \mathcal{L}_{1,i+1} \\
      \vdots \\
      \log \mathcal{L}_{T,i} - \log \mathcal{L}_{T,i+1}
    \end{bmatrix}.
    \]
    \State Compute the Jacobian of the softmax function:
    \[
    (J_i)_{tt'} = \frac{\partial z_{t,i}}{\partial \xi_{t',i}} = z_{t,i} (\delta_{tt'} - z_{t',i}).
    \]
    \State Aggregate the gradient by the chain rule:
    \[
    \delta_i = J_i^\top\, d_i.
    \]
    \State Update logits:
    \[
    \xi_{i+1} = \xi_i - \beta \bigl( \delta_i + \gamma\, \xi_i \bigr).
    \]
\EndFor
\end{algorithmic}
\end{algorithm}

\subsection{MTL-LoRA}
MTL-LoRA (Multi-Task Learning LoRA) is a parameter-efficient fine-tuning method designed to enhance the multi-task learning (MTL) capabilities of large language models (LLMs). It builds upon the Low-Rank Adaptation (LoRA) framework by addressing the challenges of task interference and suboptimal information sharing in multi-task scenarios.

LoRA is a parameter-efficient fine-tuning method that freezes the majority of a pre-trained model's parameters and introduces trainable low-rank matrices to approximate gradient updates. For a weight matrix $\mathbf{W} \in \mathbb{R}^{d \times k}$ in the original model, LoRA decomposes the gradient update $\Delta W$ into two low-rank matrices $\mathbf{B} \in \mathbb{R}^{d \times r}$ and $\mathbf{A} \in \mathbb{R}^{r \times k}$, where $r \ll \min(d, k)$. The updated weight matrix is expressed as:
\[
\mathbf{W}' = \mathbf{W} + \Delta \mathbf{W} = \mathbf{W} + \mathbf{BA}.
\]

The output of the updated layer for an input $x$ is:
\[
\mathbf{h} = (\mathbf{W} + \mathbf{BA})\mathbf{x}.
\]

MTL-LoRA enhances LoRA by introducing task-specific transformations and dynamic information-sharing strategies.

\paragraph{Task-Specific Transformation.} 
MTL-LoRA introduces a learnable task-specific transformation matrix $\mathbf{\Lambda}_t \in \mathbb{R}^{r \times r}$ for each task $t$. For an input $\mathbf{x}_t$ corresponding to task $t$, the low-rank projection is modified as:
\[
\mathbf{z}_t = \mathbf{\Lambda}_t \mathbf{A} \mathbf{x}_t,
\]
where $\mathbf{A} \in \mathbb{R}^{r \times k}$ is the shared low-rank matrix.

\paragraph{Dynamic Information Sharing.}
To improve cross-task information sharing, MTL-LoRA employs multiple up-projection matrices $\mathbf{B}_i \in \mathbb{R}^{d \times r}$ ($i = 1, \dots, n$) and combines their outputs using a weighted averaging strategy. The final output for task $t$ is computed as:
\[
\mathbf{h}_t = \mathbf{W} \mathbf{x}_t + \sum_{i=1}^{n} \frac{\exp(w_i^t / \tau)}{\sum_{j=1}^{n} \exp(w_j^t / \tau)} \mathbf{B}_i \mathbf{z}_t,
\]
where $w_i^t$ are learnable weights for task $t$, and $\tau$ is a temperature hyperparameter controlling the sharpness of the weight distribution.

We set number of up-projection matrices $n$ to 3, rank to 16 and temperature $\tau$ to 0.5

\section{LoRA Rank}
\label{app:lorarank}
\begin{figure}[t]
    \centering
    \includegraphics[width=0.5\textwidth]{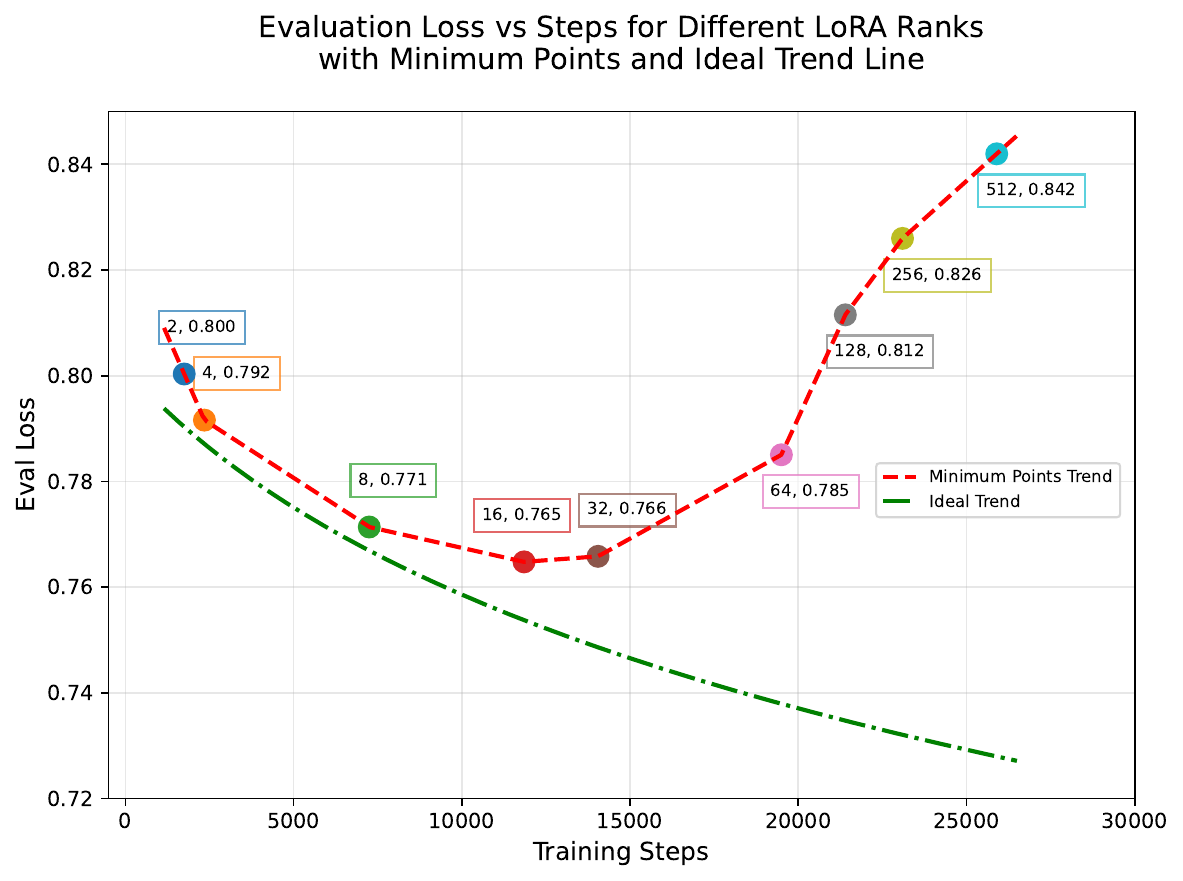} 
    \caption{The loss curve with different LoRA rank. }
    \label{fig:rankloss}
\end{figure}  
% \zz{This convergence curve will also cause lots of troubles. Instead, please use a table to show the accuracy or evaluation loss under various ranks.}
We investigate the influence of LoRA rank on the model's final performance. Initially, we exclude weight masking and fine-tune the model with different LoRA ranks. The evaluation loss curves for ranks ranging from $1$ to $512$ are plotted in Figure~\ref{fig:rankloss}. As the rank increases, the loss forms a U-shaped curve, with the lowest point occurring at a rank of $16$. Ideally, the trend of the lowest point in first-order (FO) optimization should follow the green dashed line in Figure~\ref{fig:rankloss}. However, in the zeroth-order (ZO) setting, the larger parameter optimization space as rank increases leads to a deviation from this ideal trend. 

This U-shaped curve highlights a critical trade-off: while increasing the rank improves the model's capacity, it simultaneously introduces challenges in optimizing a larger parameter space under ZO settings. This observation directly motivates our exploration of sparsity and mask selection strategies, which aim to reduce the number of parameters being optimized while retaining the most important ones. By identifying and focusing on the most critical parameters, we can mitigate the challenges posed by ZO optimization and achieve better performance, as demonstrated by our MaZO approach.

\section{Memory Usage and Search Time}
\label{sec:memory usage}

To evaluate the efficiency of MaZO, we compare its memory usage, search time, and training time against baseline vanilla multi-task learning ZO methods, both with and without LoRA. Table~\ref{tab:memory_time_comparison} summarizes the results.

\begin{table}[t]
\centering
\resizebox{\columnwidth}{!}{
\begin{tabular}{l|c|c|c}
\hline
\textbf{Method}             & \textbf{Memory (GB)} & \textbf{Search Time (min)} & \textbf{Training Time (h)} \\ \hline
MTL-ZO        &            29.0              &             -              &                  14.3           \\ 
MaZO          &            33.3               &             42              &                 16.6            \\ 
$\text{MTL-ZO}_{\text{LoRA}}$             &            31.2               &            -               &      13.7                       \\ 
$\text{MaZO}_{\text{LoRA}}$                   &            33.9               &        8.5                   &           14.1                  \\ \hline
\end{tabular}}
\caption{Comparison of memory usage, search time, and training time between MTL-ZO and MaZO, with and without LoRA. \textit{MTL} refers to multi-task learning. While MaZO introduces marginal memory and runtime overhead due to the mask storage and search process, it achieves significantly better accuracy as shown in Tables 1 and 2, demonstrating its effectiveness and practicality. Note that the memory requirement exceeds the model size (7B) because we use a batch size of 16 and a maximum token length of 600. }
\vspace{-10pt} 
\label{tab:memory_time_comparison}
\end{table}
% \zz{This table can mislead the reviewers, and they may think that our method does not have benefit. Consider two options: (1) delete this table and explain in the body text that our method achieves much better convergence with negligible memory and run-time overhead. (2) keep this table but emphasize in the caption that our method has marginal memory and run-time overhead, while achieving much better accuracy.}

The search time introduced by MaZO is negligible compared to the overall training time.
MaZO incurs a slight increase in memory usage (approximately $10\%$) compared to baseline multi-task learning ZO methods. This is primarily due to the additional storage required for the weight update mask. However, this increase is marginal and does not significantly impact the overall memory efficiency, especially when combined with LoRA, where the parameter space is already reduced.
While MaZO introduces a small memory overhead, its benefits in terms of faster convergence and reduced gradient variance outweigh this cost, making it an effective and practical solution for multi-task fine-tuning under ZO optimization.

% \section{Hyperparameter Search for Sparsity}
% \label{app:sparsity}

\section{Details of Different Weight Score Metrics}
\label{app:metrics}

\subsection{Wanda: Pruning by Weights and Activations}

In this section, we introduce \textbf{Wanda} (Pruning by Weights and Activations), a simple yet effective method for pruning large language models (LLMs). Wanda can induce high sparsity in pretrained LLMs without requiring retraining or weight updates, making it computationally efficient and easy to implement.

The key idea of Wanda is to evaluate the importance of each weight based on both its magnitude and the corresponding input activation. Specifically, for a linear layer with weight matrix $\mathbf{W} \in \mathbb{R}^{C_{\text{out}} \times C_{\text{in}}}$ and input activations $\mathbf{X} \in \mathbb{R}^{(N \cdot L) \times C_{\text{in}}}$, the importance score $\mathbf{S}_{ij}$ of weight $\mathbf{W}_{ij}$ is defined as:
\begin{equation}
    \mathbf{S}_{ij} = |\mathbf{W}_{ij}| \cdot \|\mathbf{X}_j\|_2,
\end{equation}
where $|\mathbf{W}_{ij}|$ is the absolute value of the weight, and $\|\mathbf{X}_j\|_2$ is the $\mathcal{L}_2$ norm of the $j$-th column of $\mathbf{X}$, aggregated across all tokens in the batch and sequence dimensions. This metric effectively combines weight magnitude and input activation information to determine the importance of each weight.

Unlike traditional pruning methods that compare weights globally or layer-wise, Wanda adopts a per-output comparison strategy (the same as our row-wise comparison). For a weight $\mathbf{W}_{ij}$ connecting input $j$ to output $i$, its comparison group is defined as all weights connected to the same output $i$:
\begin{equation}
    \mathbf{G}_{ij} = \{\mathbf{W}_{uv} \,|\, u = i\}.
\end{equation}
Within each comparison group, weights are ranked by their importance scores $\mathbf{S}_{ij}$, and a predefined sparsity ratio $s\%$ is applied to prune the lowest-ranked weights.

\subsection{Other Metrics}

In this section, we introduce two additional heuristic weight importance metrics: random and magnitude.

For the random metric, we randomly select $50\%$ of the weights. It is important to note that the comparison group is the entire set of weights, rather than a single row.

For the magnitude metric, we select weights with the smallest values in a weight, following the approach described by \citet{liu2024sparse}.

\section{Task Details}
\label{appendix:task_details}

We consider a diverse set of natural language understanding (NLU) and natural language generation (NLG) tasks. 

\subsection{Natural Language Understanding Tasks}
We select tasks from the GLUE \citep{wang-etal-2018-glue} and SuperGLUE \citep{wang2019superglue} benchmarks:
\begin{itemize}
    \item \textbf{SST-2} (Stanford Sentiment Treebank): A binary sentiment classification task.
    \item \textbf{BoolQ}: A yes/no question-answering task.
    \item \textbf{RTE} (Recognizing Textual Entailment): A binary classification task for textual entailment.
    \item \textbf{WSC} (Winograd Schema Challenge): A pronoun resolution task.
    \item \textbf{WiC} (Word-in-Context): A word sense disambiguation task.
    \item \textbf{MultiRC} (Multi-Sentence Reading Comprehension): A question-answering task where each question has multiple correct answers.
    \item \textbf{COPA} (Choice of Plausible Alternatives): A multiple-choice task for causal reasoning.
\end{itemize}

\subsection{Natural Language Generation Task}
For natural language generation, we include:
\begin{itemize}
    \item \textbf{SQuAD} \citep{rajpurkar2016squad}: A question-answering dataset where the model generates text-based answers from a given passage.
\end{itemize}

\subsection{Dataset Splits and Evaluation Metrics}
To ensure computational feasibility, we randomly sample 500 instances for training, 250 for validation, and 500 for testing for each task. Performance is measured using F1 score or accuracy, depending on the task.

\section{Discussion of Collinearity}
\label{app:collinear}
\textbf{Task-specific ZO gradients:} For each task \( t \in \{1,\dots,T\} \), the zeroth-order gradient estimate is given by
\begin{equation}
    \mathbf{g}^t = \frac{\mathcal{L}^t(\theta+\epsilon \mathbf{z}) - \mathcal{L}^t(\theta-\epsilon \mathbf{z})}{2\epsilon} \, \mathbf{z} \equiv \alpha_t \mathbf{z},
\end{equation}
where \(\alpha_t\) is a scalar. Thus, every \(\mathbf{g}^t\) is a scalar multiple of the same random direction \(\mathbf{z}\).

\textbf{Span of all task gradients:} The space spanned by the set of all task gradients is
\begin{equation}
    \operatorname{span}\{\mathbf{g}^1, \mathbf{g}^2, \dots, \mathbf{g}^T\} = \operatorname{span}\{\mathbf{z}\}.
\end{equation}
Therefore, the dimension of this span is
\begin{equation}
    \dim\left(\operatorname{span}\{\mathbf{g}^1, \mathbf{g}^2, \dots, \mathbf{g}^T\}\right) = 1.
\end{equation}

\textbf{Aggregated gradient:} The combined gradient used for the update is
\begin{equation}
    \mathbf{g} = \sum_{t=1}^T w_t \mathbf{g}^t = \left(\sum_{t=1}^T w_t \alpha_t \right) \mathbf{z},
\end{equation}
which clearly lies in the one-dimensional subspace spanned by \(\mathbf{z}\).

\textbf{Gradient covariance matrix:} Define the covariance matrix of the task gradients as
\begin{equation}
    \mathbf{C} = \sum_{t=1}^T \pi_t \left( \mathbf{g}^t - \bar{\mathbf{g}} \right)\left( \mathbf{g}^t - \bar{\mathbf{g}} \right)^\top,
\end{equation}
where \(\pi_t\) are probability weights (or simply \(1/T\) for uniform weighting) and the mean gradient is
\begin{equation}
    \bar{\mathbf{g}} = \sum_{t=1}^T \pi_t \mathbf{g}^t.
\end{equation}
Since \(\mathbf{g}^t = \alpha_t \mathbf{z}\), we have
\begin{equation}
    \mathbf{g}^t - \bar{\mathbf{g}} = (\alpha_t - \bar{\alpha}) \mathbf{z}, \quad \text{with } \bar{\alpha} = \sum_{t=1}^T \pi_t \alpha_t.
\end{equation}
Thus, the covariance matrix becomes
\begin{equation}
    \mathbf{C} = \left(\sum_{t=1}^T \pi_t (\alpha_t - \bar{\alpha})^2 \right) \mathbf{z} \mathbf{z}^\top.
\end{equation}
Since \(\mathbf{z}\mathbf{z}^\top\) is an outer product of a vector with itself, it has rank 1. Hence,
\begin{equation}
    \operatorname{rank}(\mathbf{C}) = 1.
\end{equation}

\textbf{Conclusion:} The lack of \emph{directional diversity} in the task gradients is mathematically captured by the fact that all task-specific gradients lie in a one-dimensional subspace, and the covariance matrix of these gradients has rank 1. This indicates that no matter how many tasks are aggregated, the update direction remains confined to a single direction \(\mathbf{z}\) in the parameter space.

\end{document}